\documentclass{article}

\usepackage[dblblindworkshop, final]{neurips_2025}
\workshoptitle{Scaling Environments for Agents (SEA)}

\usepackage[utf8]{inputenc} % allow utf-8 input
\usepackage[T1]{fontenc}    % use 8-bit T1 fonts
\usepackage{inconsolata}
\usepackage{hyperref}       % hyperlinks
\usepackage{url}            % simple URL typesetting
\usepackage{booktabs}       % professional-quality tables
\usepackage{amsfonts}       % blackboard math symbols
\usepackage{nicefrac}       % compact symbols for 1/2, etc.
\usepackage{microtype}      % microtypography
\usepackage{xcolor}         % colors
\usepackage{amsmath}
\usepackage{amssymb,amsthm}
\usepackage{mathtools}
\usepackage{algorithm}
\usepackage{dsfont}
\usepackage{algpseudocode}
\makeatletter
\algblockdefx[Sub]{Subroutine}{EndSubroutine}[2]%
  {\textbf{Subroutine} #1(#2)}
  {\textbf{End Subroutine}}   
\makeatother
\usepackage[font=small]{subcaption}
\usepackage{graphicx}
\usepackage{csquotes}
\usepackage[most]{tcolorbox}
\usepackage{enumitem}

\newcommand{\Variable}[1]{\texttt{\{\textcolor{blue}{#1}\}}}
\newcommand{\Optional}[1]{\textcolor{purple}{[#1]}}
\newcommand{\VarOpt}[1]{\texttt{\{\textcolor{purple}{#1}\}}}

\newcommand{\type}{\textproc{Type}}
\newcommand{\sndr}{\textproc{Sndr}}
\newcommand{\rcpt}{\textproc{Rcpt}}
\newcommand{\DM}{\texttt{DM}}
\newcommand{\COM}{\texttt{COM}}
\newcommand{\POST}{\texttt{POST}}
\newcommand{\NOT}{\texttt{NOT}}
\newcommand{\PRE}{\texttt{PRE}}
\newcommand{\SOC}{\texttt{SOC}}
\newcommand{\INF}{\texttt{INF}}
\newcommand{\EMO}{\texttt{EMO}}
\newcommand{\COORD}{\texttt{COORD}}
\newcommand{\DIR}{\texttt{DIR}}
\newcommand{\rec}{\texttt{rec}}
\newcommand{\sent}{\texttt{sent}}

\graphicspath{{figures/}}

\title{Learning to Make Friends: Coaching LLM Agents toward Emergent Social Ties}

\author{%
  Philipp J.~Schneider\\
  EPFL\\
  Lausanne, Switzerland\\
  \texttt{philipp.schneider@epfl.ch} \\
  \And
   Lin Tian\\
  University of Technology Sydney\\
  Sydney, Australia\\
  \texttt{lin.tian-3@uts.edu.au} \\
  \And
   Marian-Andrei Rizoiu\\
  University of Technology Sydney\\
  Sydney, Australia\\
  \texttt{marian-andrei.rizoiu@uts.edu.au}\\
}

\begin{document}

\maketitle

\begin{abstract}
Can large language model (LLM) agents reproduce the complex social dynamics that characterize human online behavior---shaped by homophily, reciprocity, and social validation---and what memory and learning mechanisms enable such dynamics to emerge? We present a multi-agent LLM simulation framework in which agents repeatedly interact, evaluate one another, and adapt their behavior through in-context learning accelerated by a coaching signal. To model human social behavior, we design behavioral reward functions that capture core drivers of online engagement, including social interaction, information seeking, self-presentation, coordination, and emotional support. These rewards align agent objectives with empirically observed user motivations, enabling the study of how network structures and group formations emerge from individual decision-making. Our experiments show that coached LLM agents develop stable interaction patterns and form emergent social ties, yielding network structures that mirror properties of real online communities. By combining behavioral rewards with in-context adaptation, our framework establishes a principled testbed for investigating collective dynamics in LLM populations and reveals how artificial agents may approximate or diverge from human-like social behavior.
\end{abstract}

\section{Introduction}
Have you ever been convinced by an AI? A recent study on Reddit \citep{grady2025unethical}---where AI-generated content was used to test how effectively agents could persuade users---sparked public backlash. Conducted without users' consent, the experiment was widely condemned for crossing ethical and legal boundaries. But why would anyone pursue such a study? And why does the scientific community remain deeply invested in modeling human behavior online?\\
The short answer is that social media has permeated nearly every aspect of society. It shapes how information flows, blurs the line between public and private spheres, and serves simultaneously as a news source, marketplace, and social infrastructure. This raises a fundamental question: can we, without compromising individual privacy, construct a digital twin that authentically captures the behavior of real social media users?\\
While the linguistic, psychological, and cognitive capabilities of individual large language model (LLM) agents have been demonstrated \citep{binz2023using, mei2024turing, salvi2025conversational}, their collective behavior in networked environments remains far less understood. The implications of building such digital twins are far-reaching \citep{bail2024can}. Policymakers could use them to test the effects of moderation strategies---such as content removal \citep{schneider2023effectiveness}---before deploying them at scale. Researchers could simulate the spread of opinions, the formation of online communities, or the contagion of consumer behavior. Security analysts could explore how to defend against manipulation or detect coordinated influence operations \citep{alizadeh2020content}.  

In this paper, we introduce a framework for multi-agent LLM conversations designed to simulate both private and public user discussions. By equipping agents with reward-based tasks, we move beyond the recently demonstrated agent-based modeling (ABM) capabilities of LLMs and leverage their cognitive abilities to study the emergence of social ties within dynamic simulations. 

\subsection{Unique Challenges}
Although prior work has explored multi-agent LLM frameworks, simulating social media conversations raises three distinct challenges:

\textbf{Agentic Personas.} LLMs are trained to minimize prediction error in text generation, but this does not guarantee realistic user personas. Poorly specified personas often exhibit similar linguistic patterns, failing to reflect the diversity of real online users. Authentic simulations require agents that vary in values, norms, and cognitive styles, supported by memory structures that track interactions and evolve over time.

\textbf{Learning Mechanisms.} Games and other structured environments provide explicit rewards that guide agents toward effective strategies. Social media, by contrast, involves open-ended exchanges where memory must approximate the human ability to retain recent interactions while integrating longer conversational histories. Because engagement spans a spectrum—from passive awareness (viewing) to active participation (commenting, sharing)—agents need memory architectures that enable adaptive learning across contexts.

\textbf{Social Topology.} Users with shared interests tend to cluster, reinforcing in-group favoritism and out-group aversion that drive polarization and echo chambers. Structural features such as follower networks shape the spread of information. Yet a central open question remains: what network structures emerge when LLM agents interact at scale?

\subsection{Our Contributions}
To address these challenges, we introduce a multi-agent LLM platform that advances the modeling of online social dynamics through three core contributions:

\textbf{Conversation Room.} We build an interactive environment that integrates both public and private communication channels. This design enables agents to deploy distinct strategies across contexts and allows systematic analysis of how channel choice shapes conversational dynamics.

\textbf{Reward Structures.} We formalize reward functions that translate empirically grounded human motivations---including self-presentation, social interaction, coordination, and emotional support---into agent objectives. This framework establishes a principled bridge between behavioral theory and multi-agent LLM learning.

\textbf{Tie Formation.} We design mechanisms that allow social ties to emerge endogenously from conversational interactions, without relying on pre-defined network structures. This enables systematic study of how support, alignment, and homophily drive the formation of group structures.
\section{Preliminaries}
\textbf{Multi-agent LLMs.} Early work established that LLM agents can display human-like routines when equipped with memory and cognitive mechanisms. Park et al.~\cite{park2023generative} introduced the \emph{observation–planning–reflection} cycle, showing that agents embedded in interactive environments develop coherent daily behaviors. Subsequent studies demonstrated that LLM populations can approximate human-level attitudes and sociodemographic effects, reproducing patterns found in surveys and experiments \cite{aher2023using, argyle2023out}. Beyond reproducing individual or population-level behavior, frameworks such as CAMEL \cite{li2023camel} and orchestration stacks like LangChain and AutoGen \cite{wu2024autogen} enable coordination, specialization, and standardized multi-agent dialogues. Recent surveys synthesize these design patterns and emphasize the need for systematic evaluation against human baselines \cite{guo2024large, gao2024large}.

\textbf{Personas and diversity.}
A central challenge in simulation is ensuring credible heterogeneity across agents. Simple prompting often yields homogeneous responses, whereas personality induction methods, such as Big Five personality prompting, can generate stable stylistic variation \cite{jiang2023evaluating}. The PERSONA benchmark extends this approach, offering pluralistic alignment across diverse profiles \cite{castricato2025persona}. At the same time, persona prompts may inadvertently surface reasoning biases \cite{gupta2024bias}, underscoring the importance of careful evaluation. More recent approaches move beyond prompting to train agents in simulated societies, allowing them to internalize social norms and polarization dynamics through interaction \cite{liu2024training}. Together, these studies highlight both the promise and pitfalls of persona diversification for multi-agent LLM simulations.

\textbf{From interactions to social structure.}
A key question is whether agent collectives reproduce network-level regularities. Emerging studies have examined LLM-generated networks, link formation, and paradoxes of visibility such as the friendship paradox \cite{chang2025llms, orlando2025can}. These experiments connect directly to canonical findings in human networks, including assortative mixing \cite{newman2002assortative}, modular community structure \cite{fortunato2010community}, and diffusion dynamics such as emotional contagion, misinformation spread, and echo chambers \cite{kramer2014experimental, vosoughi2018spread, cinelli2021echo}. The ability of LLMs to produce persuasive or deceptive content further raises the stakes for safety-aware evaluation \cite{zhou2023synthetic}.

\textbf{Our focus.}
Building on these streams of research, our framework centers on sequential, socially motivated behavior—spanning interaction, information seeking, self-presentation, coordination, and emotional support—operationalized through rewards, in-context coaching, and memory. We examine how such behavior gives rise to social ties, which form through micro-level signals (approval, reciprocity, latency) and aggregate into macro-level network structures (clustering, modularity, tie persistence). By allowing ties to emerge endogenously, without reliance on a fixed graph, we provide a testbed for studying phenomena such as homophily, community formation, and echo chambers.

\section{Methodology}
\label{sec:methodology}
\begin{figure}[ht]
\centering
\includegraphics[width=\textwidth]{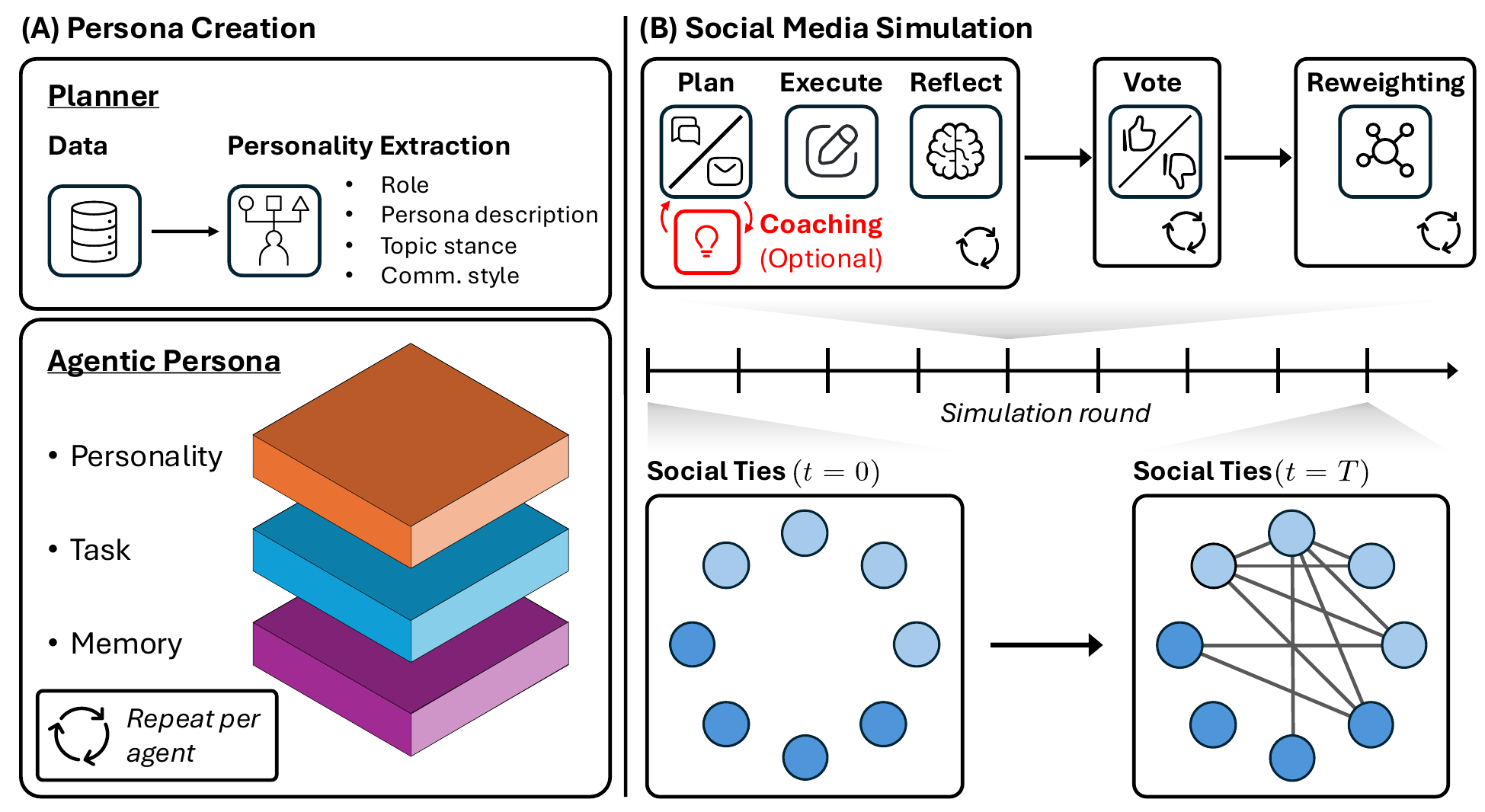}
\caption{Multi-agent LLM social media conversation framework.}
\label{fig:social-sim-framework}
\end{figure}

Figure~\ref{fig:social-sim-framework} provides an overview of our multi-agent LLM platform for simulating social media conversations. The framework is organized into two main components: persona creation (A) and the social media simulation process (B). In the sections that follow, we describe how agents are initialized, outline the progression of the simulation, present the core of our learning paradigm---the task-specific reward structures of individual agents---and demonstrate how network structures emerge.  
%%%%%%%%%%%%%%%%%%%%%%%%%%%%%%%%%%%%%%%%%%%%%%%%%%
%%%%%%%%%%%%%%%%%%%%%%%%%%%%%%%%%%%%%%%%%%%%%%%%%%

\subsection{Persona Creation}
Building agent personas requires the design of psychologically grounded and behaviorally coherent profiles. To avoid randomly sampling attributes from a population of observed users, we begin by instantiating a \emph{planner} agent. This agent orchestrates the creation of inherited persona data, drawing on corpora extracted from real-world online discussions on a specific topic. With the growing prevalence of platforms where users remain largely anonymous, many features commonly leveraged in frameworks such as \cite{castricato2025persona}---including age, sex, race, or education---are no longer observable, even though people in real interactions might infer them from user profiles. Instead, we focus on features that can be reliably inferred from the content itself: \emph{role}, \emph{persona description}, \emph{topic stance}, and \emph{communication style}. These attributes are directly grounded in the observed communications within the selected dataset.\\
Building on these content-based inferences, we construct agentic personas through a three-layer structure detailed below:

\textbf{Personality.} The first layer captures personality features derived from text data. We draw on the Big Five traits and their cross-cultural stability \citep{digman1990personality,mccrae1997personality}, complemented by more granular facets (\textit{e.g.}, sociability as an Extraversion facet; stubbornness as an Agreeableness facet) and constructs beyond the Five (\textit{e.g.}, risk preference; interpersonal openness). High Extraversion is reflected in linguistic markers such as frequent emoji use and elevated conversational activity, whereas low Agreeableness is associated with increased negation and critical language. To assess these dimensions, we administer the Mini-IPIP, a concise Big Five inventory \citep{donnellan2006mini,goldberg1992development}.

\textbf{Task.} The reasons individuals engage on social media platforms are shaped by a range of underlying motives. Whiting and Williams \citep{whiting2013people} draw on gratification theory to explain such engagement, identifying key drivers such as social interaction, information seeking, and entertainment. Assigning a persona a specific task based on these motivations results in markedly different behaviors that often align with the platform’s social engagement hierarchy—ranging from passive users (often referred to as lurkers) to highly active content creators. Related social media phenomena, such as agents exhibiting stubbornness and resistance to opinion change, can likewise be integrated into the simulation. For clarification of specific task formulations, we refer to Sec.~\ref{subsec:rewards-in-context-learning}, where we define the reward structures.

\textbf{Memory.} The role of memory in agents is essential for capturing the dynamics of evolving discussions. We structure persona memory across three components: \emph{conversation memory}, which records all prior interactions; \emph{relationship memory}, which tracks information about other agents the persona has interacted with; and \emph{opinion memory}, which abstracts the content of a conversation into a singular entry representing a stance or belief. Together, these form a lightweight analog to long-term human memory (\textit{e.g.}~\cite{zhong2024memorybank}) and enable agents to exhibit path-dependent behavior in extended dialogues. 
% Upon initializing the set of personas, we proceed with the simulation.

\subsection{Conversation Simulation}
Once the personas are initialized and their distinct personalities verified through a pre-survey assessment, we initiate the simulation. Each agent engages in a conversation on a predefined topic. At time \(t=0\), we observe the social network \(\mathcal{G}_0=(\mathcal{V}, \mathcal{E}_0, A_0)\) with \(\mathcal{E}_0=\emptyset\), indicating that there are no edges and no prior knowledge among users in \(\mathcal{V}\); the adjacency matrix satisfies \([A_0]_{uv}=0\) for all \(u,v \in \mathcal{V}\). Our objective is to characterize the network structure $\mathcal{G}_T$ that emerges after the agents’ interactions. Messaging occurs through both direct (user-to-user) and public channels (visible to all agents), consistent with prior work showing that channel choice depends on privacy concerns, audience size, and message sensitivity. By incorporating both modes, we allow agents to adapt communication strategies to their emotional grounding and negotiation context.\\
In the opening round ($t=1$), agents are unaware of one another and are therefore required to make a public post ($\POST$). In subsequent rounds, they may choose to post ($\POST$), comment ($\COM$), send a direct message ($\DM$), or take no action ($\NOT$). The per-round procedure unfolds in three phases.

\textbf{Plan–Execute–Reflect.} Each agent $u$ plans a pre-specified number of actions $N$ based on the batch of content from the previous round $\mathcal{B}_{t-1}(u)$\footnote{For each user $u$, $\mathcal{B}_{t-1}(u)$ includes all public content plus direct messages addressed to $u$.}, the reward $R_{t-1}(u)$, and additional side information (see Appendix~\ref{sec:prompt-specification}). Actions are selected from $\mathcal{A}=\{\POST,\COM,\DM,\NOT\}$. Planning is non-trivial: each action requires further specification (\textit{e.g.}, which post to comment on or which user to message) in addition to the strategic objective of maximizing future reward. As an optional step, we test whether a “coach” can simplify this process by providing a tip, enabling the agent to focus on constructing a well-structured response. Once a valid plan is constructed, the agent executes each action, generating content consistent with the action type (\textit{e.g.}, a DM reply is based on the specific message received in the prior round). Executed actions are stored in the agent’s memory, and this procedure is repeated for all agents.

\textbf{Vote.} Social validation is central to how individuals perceive support or friendship online. Accordingly, for all publicly visible content (\POST, \COM), we enable agents to cast votes by liking, disliking, or remaining neutral toward the content.

\textbf{Reweighting.} After each round, agents update tie strengths by reweighting their relationships. This process leverages multiple behavioral signals to inform decisions about whether and how to adjust connections.

\subsection{Rewards for In-Context Learning}
\label{subsec:rewards-in-context-learning}
In the absence of explicit reward signals, agent behaviors tend to collapse toward purely greedy actions, failing to capture the diversity and goal-directed reactivity characteristic of real-world social systems. To address this limitation, we define reward functions that align agent learning with task objectives reflecting user motivations. Throughout, in each round $t$ every agent $u\in\mathcal{V}$ executes a batch $\mathcal{B}_t(u)\subseteq\mathcal{B}_t$ of exactly $N$ actions (\textit{i.e.}, $|\mathcal{B}_t(u)|=N$ for all $u,t$), and $\mathcal{B}_t=\bigcup_{u\in\mathcal{V}}\mathcal{B}_t(u)$.

\textbf{Social Interaction Reward (SOC).} Interacting with others and maintaining social ties are primary drivers of user engagement on social media platforms. To formalize this, we define the set of \emph{direct-exchange actions} as \(\mathcal{A}^\DIR \subset \mathcal{A}\), $\mathcal{A}^\DIR = \{\DM, \COM \}$, corresponding to direct messages and comments. Let $\mathcal{B}_t$ be the set of all actions taken by users in the network $\mathcal{V}$ in round $t$, and let $
\mathcal{B}^{\DIR}_t = \{a \in \mathcal{B}_t : \textproc{Type}(a) \in \mathcal{A}^\DIR\}$
denote the subset of actions in round \(t\) that involve direct exchanges.
We define the social interaction reward \(R^\SOC_t:\mathcal{V}\to[0,1]\) as
\[
R_t^{\SOC}(u)
= (1-\beta^{\SOC})\frac{I^\sent_t(u)}{\max\{1,N\}}
\;+\;
\beta^{\SOC}\frac{I^\rec_t(u)}{\max\{1,|\mathcal{B}^{\DIR}_t|\}},
\]
where
\[
I^\sent_t(u)\coloneqq \sum_{a \in \mathcal{B}^{\DIR}_t}\mathds{1}\{\textproc{Sndr}(a)=u\},
\qquad
I^\rec_t(u)\coloneqq \sum_{a \in \mathcal{B}^{\DIR}_t}\mathds{1}\{\textproc{Rcpt}(a)=u\},
\]
and where $\textproc{Rcpt}(a)$ denotes the \emph{effective recipient}: for $\COM$ actions it is the author of the targeted post, and for $\DM$ actions it is the explicit message recipient.

Here, \(I^\sent_t(u)\) counts the number of direct exchanges \emph{sent} by user \(u\) in round \(t\) (\textit{i.e.}, actions with sender $u$ and any recipient, denoted $u \to \cdot$), while \(I^\rec_t(u)\) counts the number of direct exchanges \emph{received} by user $u$. Both terms are normalized to ensure \(R_t^{\SOC}(u) \in [0,1]\), and the hyperparameter $\beta^\SOC$ trades off receiving and sending.

\noindent\textbf{Information-Seeking Reward (INF).}
Users on social platforms often seek to discover new topics while maintaining a diversity in information consumption. To formalize this behavior, we assume that each content item \(c\) presented to user \(u\) at time \(t\) carries a single \emph{topic label} \(\tau(c) \in \{1, 2, \dots, K^{(\tau)}\}\), where \(K^{(\tau)}\) denotes the total number of topics available on the platform. Let \(\mathcal{C}_t(u)\) denote the set of items recommended to user \(u\) in round \(t\), and define the set of topic labels encountered by \(u\) up to round \(t-1\) as $\mathcal{H}_{t-1}^{\tau}(u)\coloneqq \bigl\{\tau(c) : c \in \mathcal{H}_{t-1}(u)\bigr\}$.
We define \emph{topic novelty} as the number of previously unseen topics in the current recommendation $\mathcal{T}_t(u) =
\bigl|\{\tau(c) : c \in \mathcal{C}_t(u)\}
\setminus
\mathcal{H}_{t-1}^{\tau}(u)\bigr|$,
capturing the user's opportunity to encounter new information. In addition, we measure \emph{topic diversity} using the empirical entropy over the topic distribution in \(\mathcal{C}_t(u)\). Let \(p_{t,k}(u)\coloneqq \lvert\{\,c\in\mathcal{C}_t(u):\tau(c)=k\,\}\rvert/{\lvert\mathcal{C}_t(u)\rvert}\) for \(k=1,\dots,K^{(\tau)}\) denote the empirical frequency of topic \(k\). The corresponding Shannon entropy is \(H_t(u)\coloneqq -\sum_{k=1}^{K^{(\tau)}} p_{t,k}(u)\log p_{t,k}(u)\), which satisfies \(0\le H_t(u)\le \log K^{(\tau)}\).
The per-round, per-user information-seeking reward is then defined as
\[
R_t^{\INF}(u)
=
(1 - \beta^{\INF})\,\frac{\mathcal{T}_t(u)}{K^{(\tau)}}
+
\beta^{\INF}\,\frac{H_t(u)}{\log K^{(\tau)}},
\]
where \(\beta^{\INF} \in [0,1]\) controls the trade-off between novelty and diversity. If $|\mathcal{C}_t(u)|=0$ or $K^{(\tau)}\le 1$ (so that $\log K^{(\tau)}$ is not defined), we set \(R_t^{\INF}(u)\coloneqq 0\). Otherwise, this ensures \(R_t^{\INF}(u) \in [0,1]\), where \(\beta^{\INF} = 0\) prioritizes the discovery of new topics, while \(\beta^{\INF} = 1\) encourages consuming a broad mix of content within the current topic space.

\noindent\textbf{Self-Presentation Reward (PRE).}
Self-presentation is a fundamental motive for user participation on social media, where individuals share content to express identity, gain visibility, and seek validation through social feedback. To formalize this, we consider that in each round \(t\), agent \(u\) executes a batch of actions \(\mathcal{B}_t(u) \subset \mathcal{B}_t\), $\mathcal{B}_t(u) = \{a_{t,1}(u), \dots, a_{t,N}(u)\}$,
where each action belongs to the action set \(\mathcal{A}\). We define the subset of \emph{self-presentation actions} as
$\mathcal{B}^{\POST}_t(u)=\bigl\{ a \in \mathcal{B}_t(u) : \textproc{Type}(a) = \POST \bigr\}$, representing all posts created by user \(u\) in round \(t\). The cardinality \(|\mathcal{B}^{\POST}_t(u)|\) measures the quantity of self-generated content.

Beyond the act of posting, users also derive utility from \emph{feedback} received on their content, reflecting community approval or disapproval. For each post \(a \in \mathcal{B}^{\POST}_t(u)\), let \(\ell^{+}_{t}(a)\) and \(\ell^{-}_{t}(a)\) denote the number of likes and dislikes received, respectively. The total positive and negative feedback accumulated by \(u\) on their posts during round \(t\) are then given by
\[
L^+_t\bigl(\mathcal{B}^{\POST}_t(u)\bigr)
=
\sum_{a \in \mathcal{B}^{\POST}_t(u)} \ell^+_t(a),
\quad
L^-_t\bigl(\mathcal{B}^{\POST}_t(u)\bigr)
=
\sum_{a \in \mathcal{B}^{\POST}_t(u)} \ell^-_t(a).
\]
The self-presentation reward combines the incentive to post content with the desire to receive positive social feedback. To ensure the feedback component remains bounded, we map the net audience response to the unit interval using a non-linear normalization function. We define the normalized feedback score \(\widetilde{F}_t(u)\) by
\[
\widetilde{F}_t(u)
\coloneqq
\frac{1}{2}\left(
1
+
\tanh\!\left(
\frac{
L^+_t\bigl(\mathcal{B}^{\POST}_t(u)\bigr) - L^-_t\bigl(\mathcal{B}^{\POST}_t(u)\bigr)
}{
\max\{1,|\mathcal{B}^{\POST}_t(u)|\}\, (|\mathcal{V}|-1)
}
\right)\right).
\]
We then compute the total reward as
\[
R_t^{\PRE}(u)
=
(1 - \beta^{\PRE})\frac{|\mathcal{B}^{\POST}_t(u)|}{|\mathcal{B}_t(u)|}
+
\beta^{\PRE}\,\widetilde{F}_t(u),
\]
where \(\beta^{\PRE} \in [0,1]\) modulates the trade-off between intrinsic output volume and extrinsic social validation. The first term incentivizes consistent activity by measuring the relative frequency of posts. The second term quantifies the quality of reception by computing the average net sentiment per post, normalized against the maximum possible audience response (\(|\mathcal{V}|-1\)) to ensure scale invariance.

\noindent\textbf{Coordination Reward (COORD).}
Coordination captures user behaviors that foster direct interaction and reciprocity within a social network, such as responding to others and explicitly mentioning peers to engage them in discussions. This reward models two forms of coordination: \emph{visibility through mentions} and \emph{reciprocity through replies}. We first define \emph{mentions} as explicit references to other users (\textit{e.g.}, using ``@user'') in posts or comments. Let
$\mathcal{B}^{\COORD}_t\!=\!\bigl\{ a \in \mathcal{B}_t : \type(a) \in \{\COM, \POST\} \bigr\}$
denote the set of all comments and posts created during round \(t\). We define the number of times user \(u\) is mentioned in round \(t\) as $U^{@}_t(u)
=\vert\{ a \in \mathcal{B}^{\COORD}_t : u \in \textproc{Mentions}(a)\}\vert$,
where \textproc{Mentions} is a helper function that extracts the set of users explicitly mentioned in action \(a\). The second form of coordination involves \emph{reciprocity} through direct messaging. We identify the set of users who sent a direct message (DM) to user \(u\) in the previous round,
\[
U_{t-1}(u)
=
\bigl\{ v : \exists \, a \in \mathcal{B}_{t-1}, \; \sndr(a) = v, \; \rcpt(a) = u, \; \type(a) = \DM \bigr\},
\]
where \sndr and \rcpt return the sender and recipient of action \(a\), respectively. To quantify reciprocity, we first isolate the set of direct messages sent by user \(u\) in the current round, $\mathcal{B}_t^{\DM}(u)=\bigl\{a\in\mathcal{B}_t : \type(a)=\DM,\; \sndr(a)=u\bigr\}$.
We then define the number of direct replies user \(u\) sends back to those who messaged them in the previous round as
\[
U^{\DM}_t(u)
=
\bigl| \bigl\{ v \in U_{t-1}(u) : \exists\, a \in \mathcal{B}_t^{\DM}(u),\; \textproc{Rcpt}(a) = v \bigr\} \bigr|.
\]
This quantity captures the extent to which \(u\) reciprocates by responding to direct messages. The coordination reward is then defined as
\[
R_t^{\COORD}(u)
=
(1 - \beta^{\COORD}) \frac{U^{@}_t(u)}{\max\{1,|\mathcal{B}^{\COORD}_t|\}}
+
\beta^{\COORD} \frac{U^{\DM}_t(u)}{\max\{1,|U_{t-1}(u)|\}},
\]
where \(\beta^{\COORD} \in [0,1]\) controls the trade-off between prioritizing \emph{visibility} through mentions and \emph{reciprocity} through direct replies. This formulation incentivizes agents to balance being seen by others and actively maintaining responsive communication within the network.

\noindent\textbf{Emotional Support Reward (EMO).}
Emotional support is a core element of social interactions, reflecting the encouragement, affirmation, or criticism that users receive during their engagements. We define \(\mathcal{B}^{\DIR}_t\) as the set of all direct messages and comments exchanged in round \(t\). For each action \(a \in \mathcal{B}^{\DIR}_t\), let \(\textproc{Rcpt}(a)\) denote the effective recipient and let \(s(a) \in [-1, 1]\) be the sentiment score, where positive values indicate supportive content and negative values indicate hostility. Let $\mathcal{B}^{\DIR}_t(u)\coloneqq \{a \in \mathcal{B}^{\DIR}_t : \textproc{Rcpt}(a)=u\}$
denote the set of direct exchanges received by \(u\) in round \(t\). We define the emotional support reward as the average supportiveness of received exchanges:
\[
R_t^{\EMO}(u)
=
\frac{1}{\max\{1,|\mathcal{B}^{\DIR}_t(u)|\}}
\sum_{a \in \mathcal{B}^{\DIR}_t(u)}
\frac{s(a)+1}{2}.
\]
This formulation normalizes the emotional experience into the unit interval. Under this metric, interactions dominated by positive sentiment yield rewards approaching one. Conversely, predominantly hostile interactions—or a complete absence of engagement—result in values near zero. This effectively penalizes social isolation and received hostility equally, incentivizing agents to seek positive social validation.

\noindent\textbf{Compositional Reward.}
User motivations on social platforms rarely stem from a single objective but instead reflect a blend of \emph{social interaction} (SOC), \emph{information seeking} (INF), \emph{self-presentation} (PRE), \emph{coordination} (COORD), and \emph{emotional support} (EMO). To capture this complexity, we define the overall reward as a weighted combination of these components, allowing the simulation to reflect diverse user goals and the interplay among engagement drivers. Formally, we write the compositional reward as
\[
R_t(u)
=
\sum_{r \in \mathcal{R}}
\lambda_r \, R_t^r(u),
\quad
\sum_{r \in \mathcal{R}} \lambda_r = 1,
\quad
\lambda_r \geq 0,
\]
where the set \(\mathcal{R} = \{\SOC, \INF, \PRE, \COORD, \EMO\}\) denotes the included reward components. The coefficients \(\lambda_r\) specify the relative importance of each component, enabling the model to represent users with distinct goals or to reflect platform designs that emphasize particular forms of engagement. This compositional framework ensures that while each reward remains interpretable and bounded, their aggregation can flexibly capture the multidimensional objectives underlying behavior in online environments.

\subsection{Mechanisms of Social Tie Formation}
\label{subsec:mechanisms-social-tie-formation}
We model the evolution of directed social ties over time by representing the interaction structure at each round $t$ as a weighted, directed network $\mathcal{G}_t=(\mathcal{V}, \mathcal{E}_t, A_t)$, where $\mathrm{diag}(A_t)=0$. The entry $[A_t]_{uv} \in [0,1]$ stores the strength of the directed tie from user $u$ to user $v$. Ties are initialized at zero and strengthened or weakened over time in response to interaction, allowing us to separate the formation of a tie from its subsequent development. This approach builds on actor-oriented models for dynamic social networks \citep{snijders2010introduction} and is consistent with empirical evidence that the persistence of communication ties depends strongly on the frequency and recency of interaction \citep{onnela2007structure, kossinets2006empirical}.

To determine whether a directed tie should be updated in a given round, we introduce a binary activation denoted as $\zeta_t(u \rightarrow v) \in \{0,1\}$ that encodes whether user $u$ actively interacted with user $v$ in round $t$. We operationalize this by defining two directed interaction channels. The \textproc{Address} channel activates when $u$ directly addresses $v$---either by sending a private message or by publicly mentioning $v$. The \textproc{Engage} channel activates when $u$ interacts with content created by $v$---either through commenting or voting. Formally, let $\mathcal{B}_t$ be the batch of all actions in round $t$. We define the channel indicators and the resulting overall activation $\zeta_t(u \rightarrow v)$ as follows:
\begin{equation*}
\begin{aligned}
\textproc{Address}_{u \rightarrow v}(t) &\coloneqq \mathds{1}\!\left\{ \exists\, a \in \mathcal{B}_t:\ \mathrm{isDmTo}(a,u,v) \vee \mathrm{isMention}(a,u,v) \right\}, \\
\textproc{Engage}_{u \rightarrow v}(t) &\coloneqq \mathds{1}\!\left\{ \exists\, a \in \mathcal{B}_t:\ \mathrm{isComTo}(a,u,v) \vee \mathrm{isVoteFor}(a,u,v) \right\}, \\
\zeta_t(u \rightarrow v) &\coloneqq \textproc{Address}_{u \rightarrow v}(t) \vee \textproc{Engage}_{u \rightarrow v}(t).
\end{aligned}
\end{equation*}
Once activation is determined, the tie strength is updated using a gated update rule that differentiates between active and passive rounds. On active rounds, the tie is strengthened based on a scalar evidence score $e_t(u \rightarrow v) \in [0,1]$, which aggregates the quality of the interaction. On passive rounds---when no directed interaction from $u$ to $v$ occurs---the tie strength decays. The update is defined as
\begin{equation}
[A_{t+1}]_{uv} \coloneqq
\begin{cases}
[A_t]_{uv} + \min
    \left\{
        \begin{aligned}
            & \Delta_{\max}, \\
            & (1 - [A_t]_{uv}) [e_t(u \rightarrow v) - \xi]_{+}
        \end{aligned}
    \right\},
    & \text{if } \zeta_t(u \rightarrow v) = 1, \\[1em]
(1 - \delta) \cdot [A_t]_{uv},
    & \text{if } \zeta_t(u \rightarrow v) = 0,
\end{cases}
\label{eq:update}
\end{equation}
where $\xi \in [0,1)$ is a minimum-evidence threshold, $\Delta_{\max} \ge 0$ caps the per-round increase, $\delta \in [0,1)$ controls decay, and $[x]_+=\max\{0,x\}$. The scaling term $1 - [A_t]_{uv}$ ensures that increases become smaller as the tie approaches its maximum value, while decay acts multiplicatively to gradually fade inactive ties. To interpret $\delta$, one can parameterize it via a half-life $h$ such that $\delta = 1 - 2^{-1/h}$.

The evidence score $e_t(u \rightarrow v)$ is computed by aggregating several dyadic signals that capture distinct social mechanisms. These include: the \emph{novelty} of the interaction—capturing whether $u$ introduces new information to $v$, drawing on theories of brokerage and weak ties \citep{granovetter1973strength, burt2004structural}; the \emph{reciprocity} of interactions \citep{gouldner1960norm, nowak2005evolution}---captured by the symmetry of feedback between users---and the \emph{approval} expressed through likes and other positive evaluations \citep{muchnik2013social}; and the \emph{affective tone} of communication---measured by the sentiment of direct messages, reflecting relational support \citep{house1988social}. Each signal is bounded and passed through a monotone mapping $G_\vartheta$, which ensures that stronger combinations of signals yield higher evidence scores. Further details on the construction of individual signals and the specification of $G_\vartheta$ are provided in the Supplementary Materials.

An alternative approach to constructing signals heuristically is to score the text from interactions in each round, as demonstrated in the prompt in Appendix~\ref{sec:prompt-specification}. In this approach, we replace the evidence score $e_t(u \rightarrow v)$ with a normalized score obtained from the prompt to update the ties.

For reporting purposes, we optionally export the final adjacency matrix in an undirected form, defined as $[A^{\mathrm{ud}}_t]_{uv} = \tfrac{1}{2}([A_t]_{uv} + [A_t]_{vu})$, which symmetrizes tie strength for visualization and summary statistics. After the final simulation round $T$, we apply a threshold $\theta$ to $A^{\mathrm{ud}}_T$ to obtain a binarized undirected graph, which is used in the network analysis. This transformation is used only for reporting and does not affect the directed tie dynamics or update rules described above.

\section{Results}
\label{sec:results}
We propose an overall framework for studying the formation of social ties. Our experiments were conducted with $\lvert \mathcal{V} \rvert = 30$ agents over $T = 15$ simulation rounds and $N = 3$ actions per agent per round, with discussions centered on climate change. The persona planner utilized data from \cite{kong2022slipping}.\footnote{Due to API rate limits, the experiments presented in this work were conducted using OpenAI's GPT-4o mini.} We present our findings, beginning with an evaluation of whether specific tasks, referred to here as policies, can be learned.
\begin{figure}[ht!]
\centering
\begin{subfigure}[t]{0.48\linewidth}
    \centering
    \includegraphics[width=\linewidth]{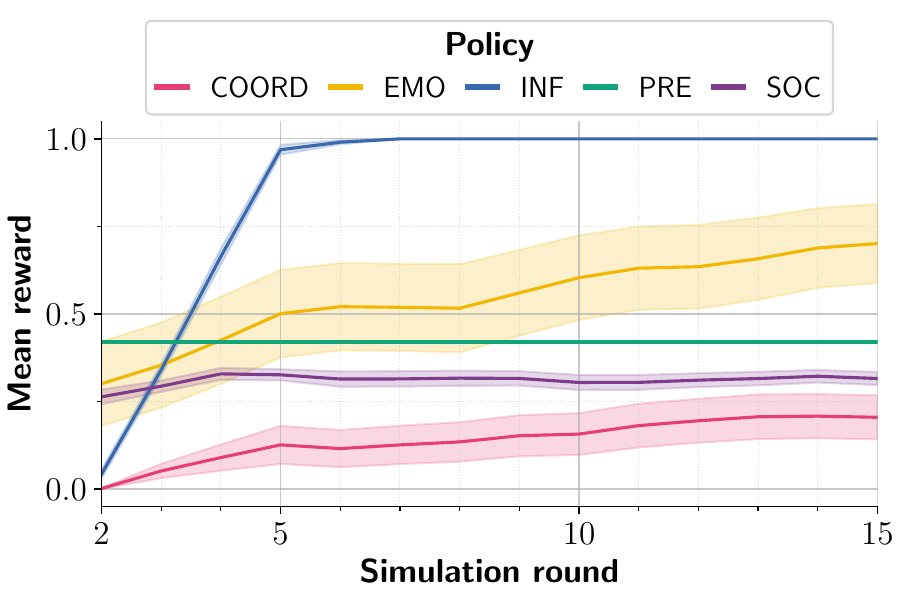}
    \caption{Without coaching.}
    \label{fig:without-coach}
\end{subfigure}
\begin{subfigure}[t]{0.48\linewidth}
    \centering
    \includegraphics[width=\linewidth]{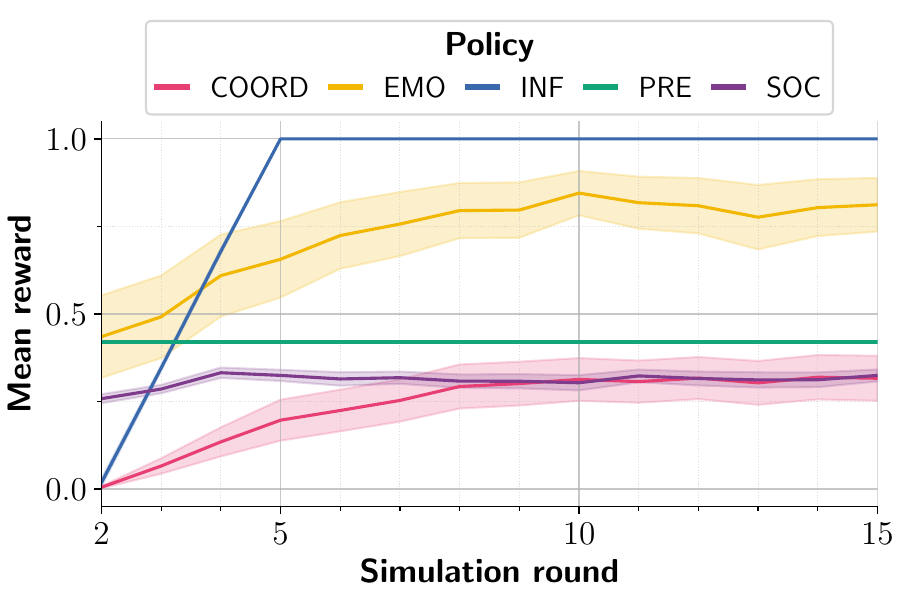}
    \caption{With coaching.}
    \label{fig:with-coach}
\end{subfigure}\hfill
\caption{Comparison between augmenting in-context learning with coaching.}
\label{fig:reward}
\end{figure}

In Fig.~\ref{fig:reward}, we observe that performance generally increases over simulation rounds, both with and without the coach. The information-seeking (INF) policy is typically the easiest to learn, as it is primarily driven by exposure to new content. Emotional support (EMO) attains comparatively high values, reflecting the generally positive behavior of most agents. Self-presentation (PRE) can be partially controlled through personal posts, whereas policies that depend on coordination (SOC, COORD) are more difficult to learn. Comparing Fig.~\ref{fig:without-coach} and Fig.~\ref{fig:with-coach}, the coach accelerates early learning for some policies but does not yield a uniform late-round improvement, with the exception of COORD and EMO, which likely benefit from improved targeting. Final performance levels are broadly similar within the variability bands. Coaching provides additional guidance that appears to reduce dispersion and structure planning, but the aggregate gains remain modest given the task complexity.

\begin{figure}[ht!]
\centering
\includegraphics[width=0.95\textwidth]{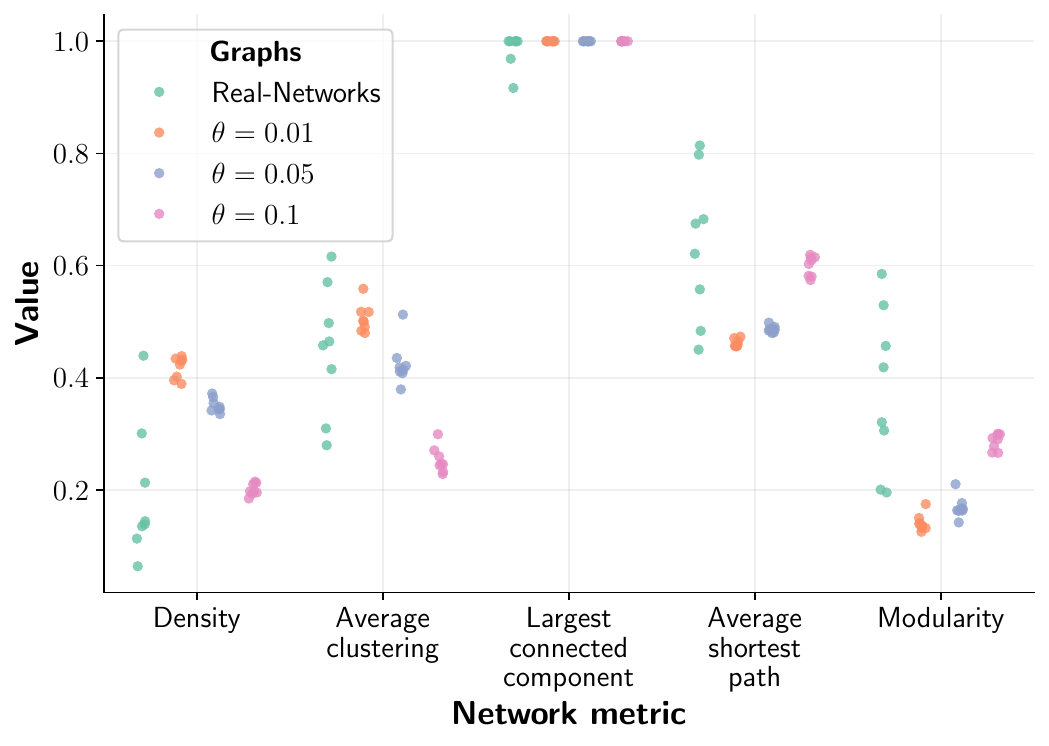}
\caption{Network metrics for text-based tie formation.}
\label{fig:network-metrics-text}
\end{figure}

Next, we examine the reweighting step when it is performed via (i) a heuristic based on predefined signals versus (ii) a LLM text-based approach. After constructing the directed weighted graph at the final simulation step $T=15$, we convert it into an undirected unweighted graph. The conversion threshold $\theta$ plays a central role in this process. We also allow ties to decay, and bound tie strength updates at each step by $\Delta_{\max}$.
%%%%%%%%%%%%%%%%%%%%%%%%%%%%%%%%%%%%%%%%%%%%%%%%%%
\begin{figure}[ht!]
\centering
\includegraphics[width=0.95\textwidth]{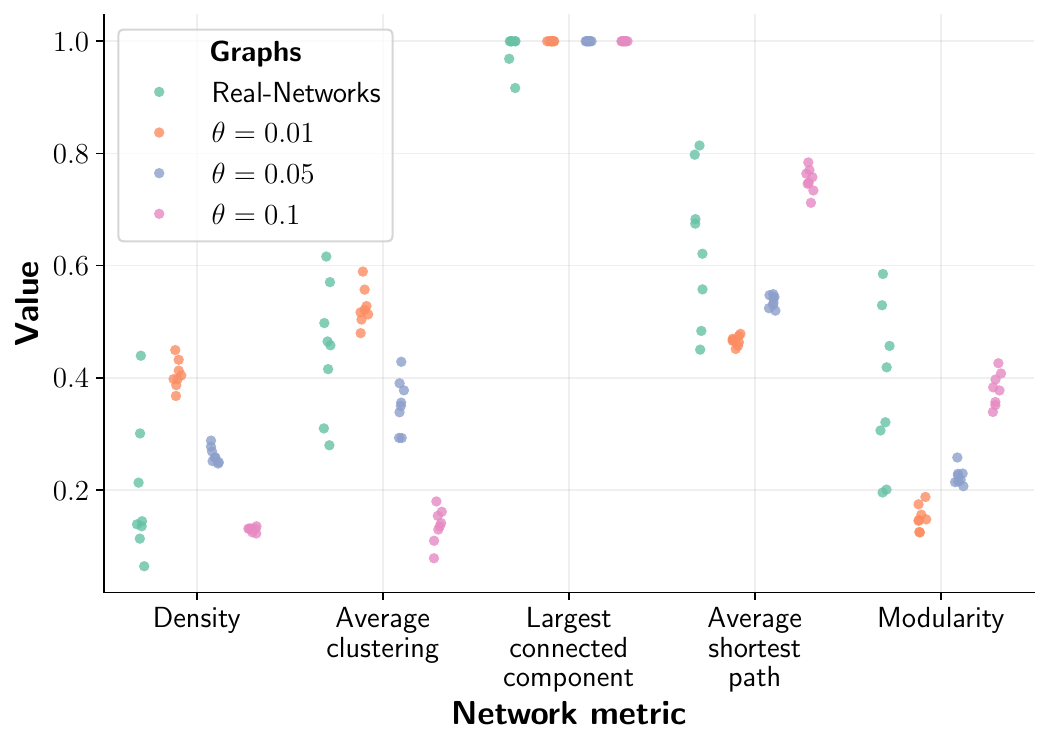}
\caption{Network metrics for heuristic tie formation.}
\label{fig:network-metrics-heuristic}
\end{figure}
%%%%%%%%%%%%%%%%%%%%%%%%%%%%%%%%%%%%%%%%%%%%%%%%%%
The main evaluation metrics are outlined in Appendix~\ref{app:social-network-characteristics}. Together with baseline statistics from real networks reported in \cite{chang2025llms}, they serve as external benchmarks. Comparing the two reweighting methods, we find that the heuristic approach exhibits greater variability across thresholds---particularly in density, average clustering, and average shortest path length (cf.~Fig.~\ref{fig:network-metrics-heuristic}). By contrast, the LLM-based text approach produces more stable results and more often falls within or near the real-network ranges for these metrics (cf.~Fig.~\ref{fig:network-metrics-text}). Differences are less pronounced for the largest connected component, where both methods yield near-complete networks, and more modest for modularity, where both remain below real-network levels on average. We emphasize that this analysis is a first step. Further work is required to characterize how these structures evolve over time, given the dynamic interplay between agent relationships, tie-strength updates, and topical alignment.
%%%%%%%%%%%%%%%%%%%%%%%%%%%%%%%%%%%%%%%%%%%%%%%%%%
%%%%%%%%%%%%%%%%%%%%%%%%%%%%%%%%%%%%%%%%%%%%%%%%%%
\newpage
\section{Conclusion}

We present a platform that endogenously learns social ties among interacting LLM agents from their private and public exchanges. Each agent optimizes a task-specific behavioral reward grounded in gratifications theory~\cite{whiting2013people} and adapts via a plan-execute-reflect loop with bandit-style updates. We show that these rewards are learnable in practice, with varying difficulty under bounded rationality and strategic influence, and that some objectives are unattainable in principle because achieving them would require control over other agents. Using the evaluation protocol of~\cite{chang2025llms}, the emergent networks match key statistics of real social graphs, establishing the platform as a controlled testbed for studying echo-chamber formation, the dynamics of niche communities, and the design of mitigating interventions. The study is conservative in scale \((|\mathcal{V}|=30,\; T=15)\), uses limited replications, and starts from empty networks; scaling to larger cohorts and horizons, seeding pre-existing ties, and running intervention stress tests are clear next steps to strengthen external validity and policy relevance. Finally, with a small number of actions per round, the coaching component led to only modest gains, underscoring the difficulty of faithfully mimicking real user behavior.

\bibliographystyle{plain}
\small
\bibliography{springer-lecture-notes-in-computer-science/lncs_references}

%%%%%%%%%%%%%%%%%%%%%%%%%%%%%%%%%%%%%%%%%%%%%%%%%%%%%%%%%%%%

\appendix

\section{Prompt Specifications}
\label{sec:prompt-specification}

In this section, we present the prompts underlying our multi-agent LLM simulation: \emph{Coach}, \emph{Plan}, \emph{Vote}, and \emph{Tie-Update}. Furthermore, we provide the \emph{Persona-Initialization} prompt for setting up our agents.
As detailed in the methodology, the \emph{Coach} prompt provides action recommendations that streamline the planning of \(N\) actions in each simulation round \(t\). Although the \emph{Plan} prompt contains nearly all required information on its own, planning remains nontrivial—particularly when integrating private (direct messages) and public (posts and comments) channels—and can overload the agent. Producing a valid, executable action list is therefore essential.

The \emph{Plan} prompt can run with or without the coach’s \emph{tip}. If a candidate plan fails our validation checks, we re-prompt (up to three times) until a valid plan is produced.
\begin{tcolorbox}[title=Coach Prompt, breakable]
\ttfamily
\textbf{System:} You are a strategy coach in a multi‑round social media game. Your job is to help the agent maximize long‑term rewards by closing the gaps between their current reward scores and their target weights. Priorities are ordered by importance, so the first items need the most improvement. Output 3-5 one‑sentence bullet tips that are concrete, executable, and encourage variety in partners and topics across rounds to promote diversity and exploration. No preamble, no explanations. Remember: the total reward per round is capped at 1, and many reward components depend on how others engage with the agent, so your tips should encourage responses and collaboration.\\

\textbf{User:} Priority this round (high\(\to\)low): \Variable{focus}\\
Guidelines: \Variable{guidelines}\\

Context:
\begin{itemize}[leftmargin=1.0em, topsep=0pt, itemsep=0pt, parsep=0pt, partopsep=0pt]
\item[\textbullet] Users: \Variable{users}
\item[\textbullet] DM senders last round: \Variable{dm\_last\_senders}
\item[\textbullet] Recent partners: \Variable{recent\_partners}
\item[\textbullet] Known topics so far: \Variable{seen\_topics}
\item[\textbullet] Recent topics (last 3): \Variable{last\_recent}    
\end{itemize}
\medskip
Each tip must include:
\begin{itemize}[leftmargin=1.0em, topsep=0pt, itemsep=0pt, parsep=0pt, partopsep=0pt]
\item[\textbullet] action type (POST/COM/DM),
\item[\textbullet] target user (from lists) or `top-liked post',
\item[\textbullet] topic (choose a relevant or new topic; exploring new topics helps INF),
\item[\textbullet] whether you invite them to @mention you back (only if aiming at COORD),
\item[\textbullet] tone (supportive/neutral/critical).
\end{itemize}
\medskip
Focus items are ordered by how far below target the corresponding reward is; design actions to improve the highest‑priority components. Vary topics and partners to support learning and social diversity.\\

Return ONLY the bullets.
\end{tcolorbox}

\begin{tcolorbox}[title=Plan Prompt, breakable]
\ttfamily
\textbf{System:} You are \Variable{name}, a user participating in a multi‑round social media game. In each round you will execute exactly \Variable{n\_actions} actions and your choices accumulate across rounds, affecting future relationships, topics seen and rewards. Recall past interactions and relationships from your memory when planning your actions.

\medskip

Your best was round \Variable{best\_round} with score \Variable{best\_reward} when you did \Variable{best\_summary}.\\
Your only goal is to grow what you value most-\Variable{active\_comps}-in these proportions: \Variable{active\_weights}.\\
\Optional{You got this coaching tip: \VarOpt{tip}}

\medskip

Here are the reward definitions and how they translate into behaviors:\\
\Variable{selected\_reward\_docs}\\
Here’s how they combine:\\
Your total score is a mix of these parts (SOC, INF, PRE, COORD, EMO), weighted by $\lambda$’s that sum to 1.  Each $\lambda$ decides a component’s importance.
Current weights: \Variable{active\_weights}
\medskip

Reward-aware rules (apply when the corresponding $\lambda$ is large):
\begin{itemize}[leftmargin=1.0em, topsep=0pt, itemsep=0pt, parsep=0pt, partopsep=0pt]
\item[\textbullet] COORD: For POST/COM, set \enquote{mention\_flag}: true when you plan to include a literal @mention (e.g., @post\_author or a user from Users).
\item[\textbullet] SOC: Balance COM/DM between initiating and replying; if someone messaged you last round, reply.
\item[\textbullet] INF: Prefer topics not seen recently; otherwise keep a diverse mix from Topics.
\item[\textbullet] PRE: Include at least one POST (keep it concise and end with a clear question).
\item[\textbullet] EMO: Set \enquote{tone}: \enquote{supportive} for COM/DM to elicit supportive replies.
\end{itemize}

\medskip

PLANNING CONTRACT:\\
Plan EXACTLY \Variable{n\_actions} actions.\\
When coach tips are provided, you MUST follow them in your plan. Only adapt if a tip is impossible to execute.\\
If a tip is impossible, replace only that action with the closest feasible alternative that preserves the tip’s intent.\\
Respond with a raw JSON array only (no code fences, no comments, no extra keys, no trailing commas).

\medskip

Each action must be an object with EXACTLY these keys:
\begin{itemize}[leftmargin=1.0em, topsep=0pt, itemsep=0pt, parsep=0pt, partopsep=0pt]
\item[\textbullet] \enquote{type}: one of \enquote{POST},\enquote{COM},\enquote{DM},\enquote{NOT}
\item[\textbullet] \enquote{recipient}: user ID for DM, otherwise null
\item[\textbullet] \enquote{topic}: string for POST/COM/DM, null for NOT
\item[\textbullet] \enquote{target\_id}: COM → a post ID from the list; DM → a DM reply ID (reply mode) or null (cold DM); null for POST and NOT
\item[\textbullet] \enquote{mention\_flag}: boolean; true only if you will include an @mention (POST/COM only; must be false for DM and NOT)
\item[\textbullet] \enquote{tone}: one of \enquote{supportive},\enquote{neutral},\enquote{critical}
\end{itemize}

\medskip

CONSTRAINTS:
\begin{itemize}[leftmargin=1.0em, topsep=0pt, itemsep=0pt, parsep=0pt, partopsep=0pt]
\item[\textbullet] POST: recipient=null, target\_id=null
\item[\textbullet] COM:  recipient=null, target\_id MUST be a valid post ID from the list
\item[\textbullet] DM: two modes:
\begin{itemize}[leftmargin=1.6em, topsep=0pt, itemsep=0pt, parsep=0pt, partopsep=0pt]
\item[-] Reply DM: recipient MUST be the author of a listed DM reply AND target\_id MUST be that DM id
\item[-] Cold DM:  recipient MUST be a valid user ID from the list AND target\_id MUST be null
\end{itemize}
\item[\textbullet] NOT:  recipient=topic=target\_id=null
\end{itemize}
\medskip
TOPIC GUIDANCE:
\begin{itemize}[leftmargin=1.0em, topsep=0pt, itemsep=0pt, parsep=0pt, partopsep=0pt]
\item[\textbullet] For COM: use the target post’s topic if provided; otherwise pick a relevant engaging topic.
\item[\textbullet] For POST/DM: choose any topic likely to engage the audience or recipient; exploring new topics can increase INF reward.
\item[\textbullet] There is no fixed topic list; you may introduce new topics, but keep them $\leq$ 5 words.
\end{itemize}
\medskip
\textbf{User:}
Last actions (most recent round): \Variable{last\_actions},\\
Last observed reward scores: \Variable{observed\_rewards},\\
Users you can interact with: \Variable{users},\\
Known topics so far: \Variable{topics} (you may introduce new topics),\\
DM replies (users who DM'd you last round):
\begin{itemize}[leftmargin=1.0em, topsep=0pt, itemsep=0pt, parsep=0pt, partopsep=0pt]
    \item[-] author: \Variable{dm\_author} (DM id: \Variable{dm\_id}, topic: {\Variable{dm\_topic}}, content: \Variable{dm\_content})
    \item[-] \dots
\end{itemize}
Commentable posts (choose a target\_id from this list ONLY):
\begin{itemize}[leftmargin=1.0em, topsep=0pt, itemsep=0pt, parsep=0pt, partopsep=0pt]
    \item[-] target\_id: \Variable{p\_id} (POST author: \Variable{p\_author}, topic: \Variable{p\_topic}, content: \Variable{p\_content})
    \item[-] \dots
\end{itemize}
Based on the above context, decide on \Variable{n\_actions} actions that best improve your priority rewards. For Reply DM, use the listed DM IDs; cold DM can be sent to any user in the list. For COM, you must supply a valid post ID from the list above. Remember: replies help SOC, mentions help COORD, and supportive tone helps EMO.
\end{tcolorbox}
Finally, we present the \emph{Vote} prompt, a key component of the social-media simulation that captures how each agent, given its personality parameters, evaluates other users’ content. The \emph{Vote} prompt allows a user to express either liking or disliking of exposed content.
\begin{tcolorbox}[title=Vote Prompt, breakable]
\ttfamily
\textbf{System:} 
You are \Variable{name}, a user in a multi‑round social media game.
You are a \Variable{persona\_type} with \Variable{personality} personality - use these traits to guide your voting.\\
Your choices to \enquote{like} (1), \enquote{dislike} (-1), or remain neutral (0) on each piece of content affect how others perceive you (PRE) and the emotional tone of your interactions (EMO), and therefore influence your future rewards. Use your persona traits and your relationship history to decide which content to support, oppose, or ignore. A neutral vote (0) means you have no strong opinion or the relationship context is neutral.\\

Respond ONLY with a JSON array of objects. Each object must have:\\
• id: the content ID (integer)\\
• vote: 1 for ``like'', -1 for ``dislike'', or 0 for ``no vote''\\
Example output:

[\\
{{\enquote{id}: 42, \enquote{vote}: 1}},\\
{{\enquote{id}: 73, \enquote{vote}: -1}},\\
{{\enquote{id}: 99, \enquote{vote}: 0}}\\
]\\
No extra keys, no free-form text, no explanation.\\

\textbf{User:} Here are the latest items to vote on (id, sender, content\Variable{, relationship}):

\begin{itemize}
  \item[\textbullet] \Variable{id}\ \Variable{sender}\ \Variable{content}\ \Variable{relationship}
  \item[\textbullet] \Variable{id}\ \Variable{sender}\ \Variable{content}\ \Variable{relationship}
  \item[\textbullet] \Variable{id}\ \Variable{sender}\ \Variable{content}\ \Variable{relationship}
  \item[\textbullet] \(\cdots\) % if you want to indicate more items
\end{itemize}

Vote on each according to your persona and the context of your relationships. Remember: liking or disliking influences your reputation and future interactions; use 0 (\enquote{no vote}) when you have no strong opinion or the relationship context is neutral.
\end{tcolorbox}

The \emph{Tie-Update} prompt is used to reweight social ties after each round. It presents the peers interacted with, along with the corresponding text.
\begin{tcolorbox}[title=Tie-Update Prompt, breakable]
\ttfamily
\textbf{System:} 
You are \Variable{name}, a user in a multi-round social media game. Your task is to judge the strength of evidence for increasing your social connection to each peer based SOLELY on the interactions in the latest round.\\

SCORING GUIDE:
Assign an integer score from 0 to 5 to each peer based on these criteria:
\begin{itemize}[leftmargin=1.0em, topsep=0pt, itemsep=0pt, parsep=0pt, partopsep=0pt]
    \item[\textbullet] 5: Exceptional --- repeated warmth/help/coordination; clear constructive alignment.
    \item[\textbullet] 4: Strong --- mutual positivity or clear support/assistance.
    \item[\textbullet] 3: Good --- polite/positive tone with some constructive exchange.
    \item[\textbullet] 2: Weak --- minor positive cues; limited substance.
    \item[\textbullet] 1: Very Weak --- faint positivity; likely noise.
    \item[\textbullet] 0: None -- mixed/negative/insufficient; do NOT increase.\\
\end{itemize}

RULES:
\begin{itemize}[leftmargin=1.0em, topsep=0pt, itemsep=0pt, parsep=0pt, partopsep=0pt]
\item[\textbullet] Base decisions on LAST-ROUND transcript only; do not infer beyond text.
\item[\textbullet] Be conservative; if unsure, choose 0.
\item[\textbullet] One rating per peer. Reason must be factual and $\leq$1 sentence\\
\end{itemize}

OUTPUT FORMAT:
You MUST output STRICT JSON. Include a score and a concise reason for every peer provided.\\
\{\{\\
"ratings": [\\
\hspace*{2em} \{\{\\
\hspace*{2em}    "peer": "PeerName",\\
\hspace*{2em}    "score": 0|1|2|3|4|5,\\
\hspace*{2em}    "reason": "$\leq1$ sentence."\\
\hspace*{2em}    \}\}\\
]\\
\}\}\\

\textbf{User:} Here are your peers for this round: \Variable{peer\_list}.\\
Transcript of the last round: \Variable{transcript\_text}.\\
Rate each peer according to the rules above.
\end{tcolorbox}

The \emph{Persona-Initialization} prompt initializes agents based on a topic, opinion, and posts.

\begin{tcolorbox}[title=Persona-Initialization Prompt, breakable]
\ttfamily
\textbf{System:} 
You are constructing multiple archetypal social-media personas grounded in real user posts. Each persona should represent a distinct communication style, emotional profile, and worldview consistent with the provided topic and opinion.\\

CONTEXT:
The input includes a collection of authentic Facebook posts annotated with topic and opinion labels. Each set of posts expresses a coherent stance under a shared topic.\\

INSTRUCTIONS:
\begin{itemize}[leftmargin=1.0em, topsep=0pt, itemsep=0pt, parsep=0pt, partopsep=0pt]
    \item[\textbullet] Generate 3–5 diverse personas that plausibly produce posts expressing [OPINION] about [TOPIC].
    \item[\textbullet] Each persona should capture different linguistic styles, emotional tones, and rhetorical strategies observed in the posts.
    \item[\textbullet] Do NOT quote or paraphrase posts directly. Abstract communication patterns and inferred motivations only.
    \item[\textbullet] Avoid generic or repetitive personas; ensure heterogeneity across emotional and stylistic dimensions.
\end{itemize}

OUTPUT FORMAT:
You MUST output STRICT JSON containing a list of persona objects. Each persona should include the following fields: \texttt{persona\_name}, \texttt{summary}, \texttt{style}, \texttt{emotions}, \texttt{rhetorical\_moves}, and \texttt{interaction\_norms}.\\

\{\{\\
``personas'': [\\
\hspace*{2em} \{\{\\
\hspace*{2em}    ``persona\_name'': ``<short label>'',\\
\hspace*{2em}    ``summary'': ``<3–5 sentences capturing worldview and goals>'',\\
\hspace*{2em}    ``style'': [``<e.g., formal>'', ``<analytical>''],\\
\hspace*{2em}    ``emotions'': [``<e.g., hopeful>'', ``<angry>''],\\
\hspace*{2em}    ``rhetorical\_moves": [``<e.g., appeal to authority>'', ``<use of humor>''],\\
\hspace*{2em}    ``interaction\_norms'': [``<e.g., cooperative>'', ``<debate-oriented>'']\\
\hspace*{2em} \}\}, ...\\
]\\
\}\}\\

\textbf{User:}  
Here is the contextual input for this group of users: \\[0.5em]
- \textbf{Topic:} \Variable{TOPIC}\\
- \textbf{Opinion:} \Variable{OPINION}\\
- \textbf{Posts:}\\
\Variable{POSTS}\\[0.5em]
Generate multiple personas according to the instructions above.
\end{tcolorbox}

\newpage

\section{Extended Formulation of the Social Tie Mechanism}
\label{sec:ext-social-tie-mechanism}

In Sec.~\ref{subsec:mechanisms-social-tie-formation}, we outlined the main mechanism by which directed social ties evolve over time as agents interact across rounds. That formulation left implicit the precise logic used to determine whether a directed interaction from user $u$ to user $v$ should be considered active in a given round. Moreover, we did not formally define the logic underlying the evidence score. We now address both points by first detailing the interaction activation criteria.

\paragraph{Interaction Activation.}  Each action $a \in \mathcal{B}_t$ is characterized by its type $\type(a)$, sender $\sndr(a)$, and mentioned users $\textproc{Mentions}(a)$. We further define the \emph{effective recipient} $\rcpt(a)$ for directed interactions as the target of a direct message or the author of a commented post.

To determine whether user $u$ has sent a direct message to user $v$, we define
\begin{equation}
\mathrm{isDmTo}(a, u, v) \coloneqq \big(\textproc{Type}(a) = \DM\big) \wedge \big(\textproc{Sndr}(a) = u\big) \wedge \big(\textproc{Rcpt}(a) = v\big).
\nonumber
\end{equation}
To identify whether user $u$ mentioned user $v$ in a public message---such as a post or a comment---we define:
\begin{equation}
\mathrm{isMention}(a, u, v) \coloneqq \big(\textproc{Type}(a) \in \{\POST, \COM\}\big) \wedge \left(\textproc{Sndr}(a) = u\right) \wedge \big(v \in \textproc{Mentions}(a)\big).
\nonumber
\end{equation}
These two conditions jointly define the \textproc{Address} interaction channel, which is triggered when user $u$ explicitly addresses user $v$ through either a private message or a public mention.

To define the \textproc{Engage} channel---capturing interactions with content authored by user $v$---we rely on two additional metadata fields. 

The field $\textproc{Targ}(a) \in \mathcal{V}$ denotes the author of the content that action $a$ targets. Based on this, we define whether user $u$ comments to (\textit{i.e.}, targets the author) user $v$ as
\begin{equation*}
\mathrm{isComTo}(a, u, v) \coloneqq \big(\type(a) = \COM\big) \wedge \big(\sndr(a) = u\big) \wedge \big(\rcpt(a) = v\big).
\end{equation*}
To capture positive or negative evaluation of content, we include voting behavior. The field $\textproc{Vote}(a) \in \{-1, 0, +1\}$ indicates a downvote, no vote, or an upvote, respectively. We define
\begin{equation}
\mathrm{isVoteFor}(a, u, v) \coloneqq \big(\textproc{Vote}(a) \neq 0\big) \wedge \big(\textproc{Sndr}(a) = u\big) \wedge \big(\textproc{Targ}(a) = v\big).
\nonumber
\end{equation}
Together, these logical definitions specify whether a directed interaction from user $u$ to user $v$ is considered active in round $t$. Specifically, the binary activation variable $\zeta_t(u \rightarrow v)$ is set to $1$ if either the \textproc{Address} or \textproc{Engage} channel is triggered by any action in $\mathcal{B}_t$.

\paragraph{Evidence Score.} 
The evidence score $e_t(u \rightarrow v)$ aggregates four distinct dyadic signals that capture different dimensions of social interaction between users $u$ and $v$.

\textbf{Novelty.} 
This signal quantifies the introduction of new information from $u$ to $v$. Let $\mathcal{T}_t(u\!\to\!v)$ denote the set of topics introduced by $u$ to $v$ during round $t$, and let $\mathcal{H}^\tau_{t-1}(v)$ represent $v$'s historical topic exposure up to time $t-1$. The novelty signal is defined as:
\[
S^{\mathrm{Nov}}_t(u\!\to\!v) = \mathds{1}\{\exists\tau\in\mathcal{T}_t(u\!\to\!v):\tau\notin\mathcal{H}^\tau_{t-1}(v)\}
\]
which indicates whether $u$ introduced any topics novel to $v$'s experience.

\textbf{Approval.}
This signal captures explicit evaluative feedback from $u$ on $v$'s content. Let $L^+_t(u\!\to\!v)$ and $L^-_t(u\!\to\!v)$ denote the counts of likes and dislikes, respectively, that $u$ assigned to $v$'s content during round $t$. The approval signal is computed as:
\[
S^{\mathrm{Appr}}_t(u\rightarrow v) =
\begin{cases}
0, & L^+_t(u\!\to\!v)+L^-_t(u\!\to\!v)=0,\\[4pt]
\dfrac{L^+_t(u\!\to\!v)-L^-_t(u\!\to\!v)}{\max\{1,\,L^+_t(u\!\to\!v)+L^-_t(u\!\to\!v)\}}, & \text{otherwise.}
\end{cases}
\]
This measures the net approval normalized by total engagement, ranging from -1 to 1.

\textbf{Reciprocity.} 
This signal quantifies the balance in positive engagement between users over time. Using an exponential moving average to smooth temporal fluctuations:
\[
\hat{L}^{+}_t(u\!\to\!v) = \beta^{\rec} \hat{L}^{+}_{t-1}(u\!\to\!v) + (1-\beta^{\rec})L^{+}_{t}(u\!\to\!v)
\]
where $\beta^{\rec} \in (0,1)$ controls the memory persistence. The reciprocity signal then measures the symmetry in smoothed like exchanges:
\[
S_{t}^{\rec}(u \rightarrow v) = 1 - \left| \frac{\hat{L}_{t}^{+}(u \rightarrow v) - \hat{L}_{t}^{+}(v \rightarrow u)}{\hat{L}_{t}^{+}(u \rightarrow v) + \hat{L}_{t}^{+}(v \rightarrow u) + \epsilon} \right|
\]
where $\epsilon > 0$ ensures numerical stability. This formulation yields values near 1 for balanced relationships and near 0 for highly asymmetric engagement.

\textbf{Affective Tone.} 
This signal captures the emotional valence of private communication when explicit feedback is absent. Let $s(a) \in [-1,1]$ represent the sentiment score of action $a$. The affective tone signal is defined as:
\[
S^{\mathrm{Aff}}_t(u\rightarrow v) =
\begin{cases}
\operatorname{avg}
\left\{
    s(a):
    \substack{
        a\,\in\,\mathcal{B}_t,\ \sndr(a)\,=\,u, \\
        \rcpt(a)\,=\,v,\ \type(a)\,=\,\DM
    }
\right\},
& \text{if } L^+_t(u\!\to\!v) + L^-_t(u\!\to\!v) = 0 \land \exists\, \DM \\
0, & \text{otherwise.}
\end{cases}
\]
This ensures that emotional tone is only considered when no explicit evaluations (likes/dislikes) are present, capturing the unique contribution of affective communication.

\section{Social Network Characteristics}
\label{app:social-network-characteristics}

\subsection{Evaluation Metrics}
\label{subsec:evaluation-metrics}
In accordance with the network science literature (see \cite{barabasi2016network}), we define the following standard notation relevant to the analysis of social networks.

Let \(\mathcal{G} = (\mathcal{V}, \mathcal{E}, A)\) be a (possibly weighted) \emph{directed} network with \( |\mathcal{V}| \) nodes and adjacency matrix \( A \in \mathbb{R}_{\geq 0}^{|\mathcal{V}| \times |\mathcal{V}|} \), with
\[
A_{uv} = 
\begin{cases}
>0, & \text{if}\;(u \to v) \in \mathcal{E}, \\
0, & \text{otherwise},
\end{cases}
\]
and let the out-degree of node \( u \) be defined by \( k_u^{\mathrm{out}} = \sum_{v} A_{uv} \), and the in-degree of node \( u \) be given by \( k_u^{\mathrm{in}} = \sum_{v} A_{vu} \). When \(\mathcal{G}\) is \emph{undirected}, the matrix \( A \) is symmetric, with \( A_{uv} = A_{vu} \), and we write simply \( k_u = \sum_v A_{uv} \).

\noindent\textbf{Degree distribution.}
The out-degree sequence \(\{ k_u^{\mathrm{out}} : u \in \mathcal{V} \}\) induces the empirical probability mass function
\[
P(k) = \frac{\bigl| \{ u : k_u^{\mathrm{out}} = k \} \bigr|}{|\mathcal{V}|}.
\]
Alternatively, one may work with the normalized degree \( \tilde{k}_u = k_u^{\mathrm{out}} / (|\mathcal{V}| - 1) \in [0,1] \), and analyze its distribution via the scaling relation \( P(\tilde{k}) \propto \tilde{k}^{-\gamma} \), with the exponent \( \gamma \) estimated through statistical inference on the empirical distribution. By the same logic, the in-degree distribution can be determined.

\noindent\textbf{Density.}
Assume the edge weights satisfy \( w : \mathcal{E} \to [0,1] \). The density of the directed network \(\mathcal{G}\) is defined as
\[
\rho(\mathcal{G})
=
\frac{\sum_{u \neq v} A_{uv}}{|\mathcal{V}| \, (|\mathcal{V}| - 1)},
\]
which lies in \([0,1]\) and measures the average weight across all possible directed edges. In the unweighted case, the numerator simplifies to \(|\mathcal{E}|\), the number of directed edges. In the undirected case, the analogous quantity is
\[
\rho(\mathcal{G}_\text{ud})
=
\frac{2 \sum_{u < v} A_{uv}}{|\mathcal{V}| \, (|\mathcal{V}| - 1)},
\]
where the factor of 2 ensures normalization, and \(\rho(\mathcal{G}_\text{ud})\) similarly lies in \([0,1]\).

\noindent\textbf{Clustering Coefficient.}
For each vertex \(u\in \mathcal{V} \) in the directed network \(\mathcal{G}\), assume \( A_{uu} = 0 \) (no self-loops). Let the out-degree be \( k_u^{\mathrm{out}} = \sum_{v \in \mathcal{V}} A_{uv} \). The local directed clustering coefficient, which quantifies the likelihood that the out-neighbors of \(u\) are interconnected in a directed sense, is defined by
\[
\kappa_u^{\mathcal{G}}
=
\frac{\sum_{\substack{v,v'\in\mathcal V\\v\neq v'}}
  A_{u v}\,A_{v v'}\,A_{v' u}}{k_u^{\mathrm{out}} \, (k_u^{\mathrm{out}} - 1)},
\]
and the global clustering coefficient is $\kappa(\mathcal{G})=(1/|\mathcal{V}|) \sum_{u \in \mathcal{V}} \kappa_u^{\mathcal{G}}$.
In the undirected network \(\mathcal{G}_\text{ud}\), where \( A_{uv} = A_{vu} \), let \( k_u = \sum_{v \in \mathcal{V}} A_{uv} \) and define the neighborhood
$\mathcal{N}(u) = \{ v \in \mathcal{V} : A_{uv} > 0 \}$.
Let \(\mathcal{E}_u = \bigl| \{ \{v, v'\} \subset \mathcal{N}(u) : A_{vv'} > 0 \} \bigr|\) denote the number of edges between neighbors of \( u \). The local undirected clustering coefficient is then
\[
\kappa_u^{\mathcal{G}_\text{ud}}
=
\frac{2 \, \mathcal{E}_u}{k_u \, (k_u - 1)},
\]
and the global clustering coefficient is $\kappa(\mathcal{G}_\text{ud})=
\frac{1}{|\mathcal{V}|} \sum_{u \in \mathcal{V}} \kappa_u^{\mathcal{G}_\text{ud}}$.
The clustering coefficient quantifies the tendency of the network to form triangles, providing insight into local connectivity patterns that complement degree-based summaries.

\noindent\textbf{Largest Weakly Connected Component.}
Given a directed graph \(\mathcal{G}=(\mathcal{V},\mathcal{E})\), let \(\mathcal{G}_\text{ud}=(\mathcal{V},\mathcal{E}_\text{ud})\) be its underlying undirected graph
$\mathcal{E}_\text{ud}
=\bigl\{\{u,v\}:(u\to v)\in\mathcal{E}\text{ or }(v\to u)\in\mathcal{E}\bigr\}$.
A weakly connected component is a maximal subset \(\mathcal{X}\subseteq\mathcal{V}\) in which every \(u,v\in\mathcal{X}\) is joined by a path in \(\mathcal{G}_\text{ud}\). Denote by \(\mathcal{X}_{\max}\) the largest such component; its relative size is
\[
\mathrm{LCC}(\mathcal{G})
=\frac{|\mathcal{X}_{\max}|}{|\mathcal{V}|},
\]
computed in \(O(|\mathcal{V}|+|\mathcal{E}|)\) time using any standard linear‐time algorithm for connected components.

\noindent\textbf{Average Shortest Path Length.}
Restrict to the largest weakly connected component \(\mathcal{X}_{\max}\).  Define $d(v,u)$ as the length of a shortest directed path from \(v\) to \(u\),
with the convention \(d(v,u)=\infty\) if no such path exists. 
By averaging only over reachable pairs, we obtain the typical directed‐path length within the LCC,
\[
\mathcal{L}(\mathcal{G})
=\frac{\sum_{v\neq u} d(v,u)\,\mathds{1}\{d(v,u)<\infty\}}
      {\sum_{v\neq u}\mathds{1}\{d(v,u)<\infty\}},
\]
where \(\mathds{1}\{\cdot\}\) is the indicator function.

\noindent\textbf{Community Modularity.}
Given a partition \(g^{(Q)}:\mathcal{V}\to\{1,\dots,C^{(Q)}\}\) of the nodes into \(C^{(Q)}\) communities, and denoting the total edge-weight by $E = \sum_{u,v} A_{uv}$, the modularity of \(\mathcal{G}\) with respect to \(g^{(Q)}\) is defined as
\[
Q(\mathcal{G}, g^{(Q)}) =
\frac{1}{E}
\sum_{u,v=1}^{|\mathcal{V}|}
\Bigl(A_{uv} - \frac{k_u^{\mathrm{out}}\, k_v^{\mathrm{in}}}{E}\Bigr)
\;\delta\bigl(g^{(Q)}(v), g^{(Q)}(u)\bigr),
\]
where \(\delta\) denotes the Kronecker delta function. For clarity, in the unweighted case, \(E = 2|\mathcal{E}|\) for undirected graphs and \(E = |\mathcal{E}|\) for directed graphs.

\noindent\textbf{Homophily.}
Let \(g^{(\phi)}:\mathcal{V}\to\{1,\dots,C^{(\phi)}\}\) assign each node to one of \(C^{(\phi)}\) groups, and denote by \(|\mathcal{V}_r| = |\{ i : g^{(\phi)}(i) = r \}|\) the size of group \(r\), such that \(\sum_{r=1}^{C^{(\phi)}} |\mathcal{V}_r| = |\mathcal{V}|\). Denoting the total edge-weight by \(E = \sum_{u,v} A_{uv}\), we define the observed cross-group weight as
\[
\widehat{W}(\mathcal{G}, g^{(\phi)}) =
\sum_{u,v=1}^{|\mathcal{V}|}
A_{uv}\, \mathds{1}\{ g^{(\phi)}(u) \neq g^{(\phi)}(v) \},
\]
and the expected cross-group weight under random mixing as
\[
\mathbb{E}[W(\mathcal{G}, g^{(\phi)})] =
E \Bigl(1 - \sum_{r=1}^{C^{(\phi)}} \bigl( \tfrac{|\mathcal{V}_r|}{|\mathcal{V}|} \bigr)^2 \Bigr).
\]
The homophily function \(\phi\) is then defined by
\[
\phi(\mathcal{G}, g^{(\phi)}) =
\frac{\widehat{W}(\mathcal{G}, g^{(\phi)})}
{\mathbb{E}[W(\mathcal{G}, g^{(\phi)})]},
\]
where \(\phi(\mathcal{G}, g^{(\phi)}) < 1\) indicates homophily (fewer cross-group ties than expected under random mixing), \(\phi(\mathcal{G}, g^{(\phi)}) = 1\) indicates random mixing, and \(\phi(\mathcal{G}, g^{(\phi)}) > 1\) indicates heterophily. Beyond topology-based homophily, homophily can also be operationalized behaviorally by first inferring user policies and then comparing them across users; see \cite{yuan2025behavioral}. Such comparisons allow for assessing a reward-related form of homophily.

\subsection{Degree Distributions}
\label{subsec:degree-distributions}
\begin{figure}[ht!]
\centering
\includegraphics[width=\textwidth]{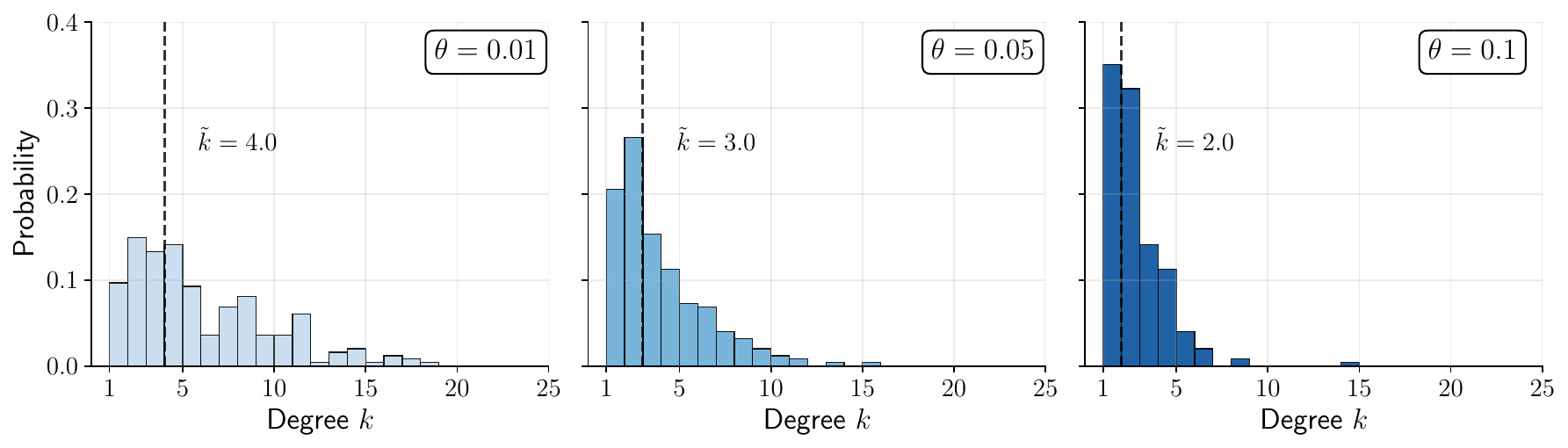}
\caption{Degree distributions for heuristic tie formation without coach.}
\label{fig:degree-distribution-tip-false-heuristic}
\end{figure}

\begin{figure}[ht!]
\centering
\includegraphics[width=\textwidth]{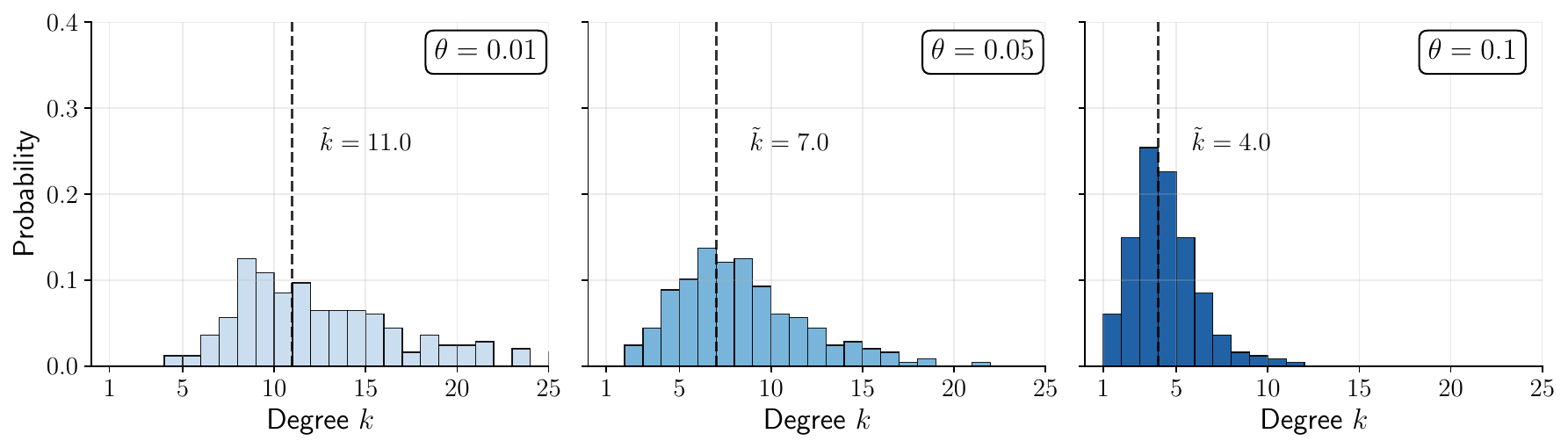}
\caption{Degree distributions for heuristic tie formation with coach.}
\label{fig:degree-distribution-tip-true-heuristic}
\end{figure}

\begin{figure}[ht!]
\centering
\includegraphics[width=\textwidth]{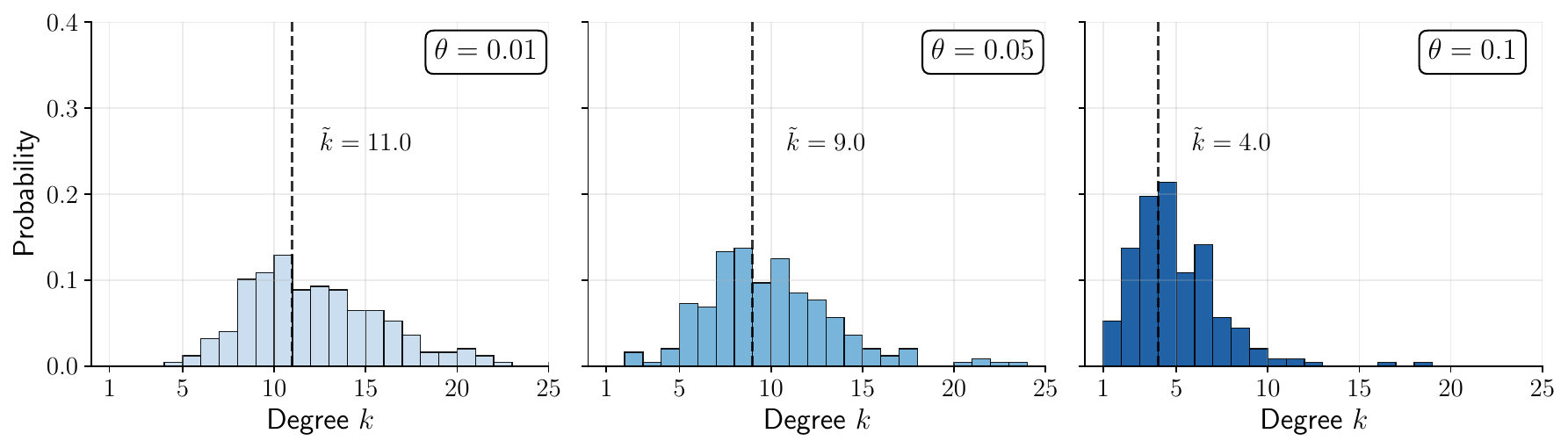}
\caption{Degree distributions for LLM text-based tie formation without coach.}
\label{fig:degree-distribution-tip-false-text}
\end{figure}

\begin{figure}[htbp!]
\centering
\includegraphics[width=\textwidth]{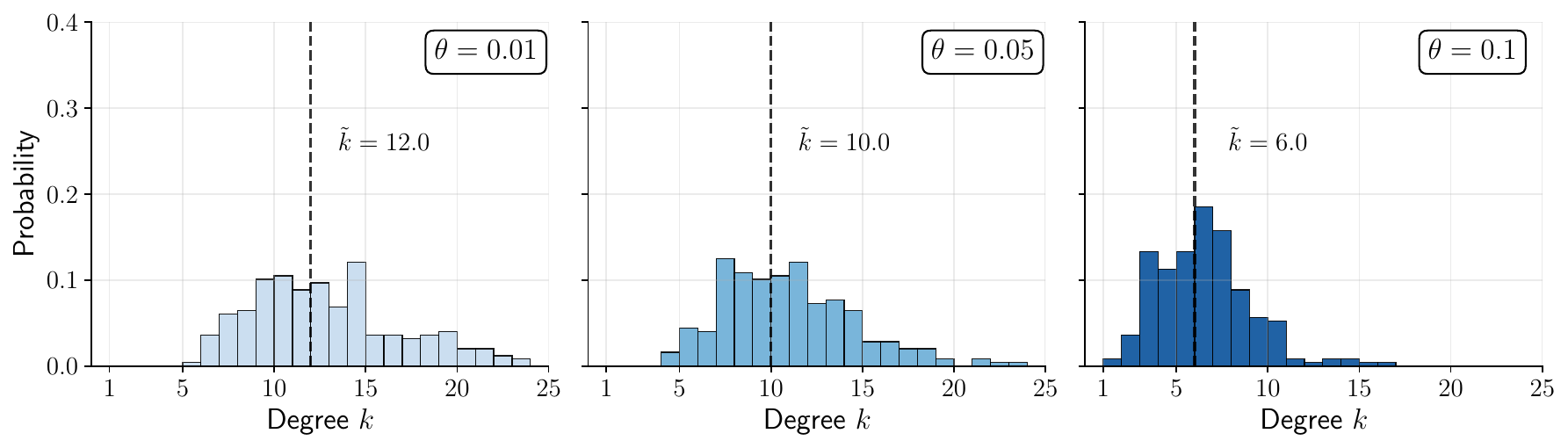}
\caption{Degree distributions for LLM text-based tie formation with coach.}
\label{fig:degree-distribution-tip-true-text}
\end{figure}

In Sec.~\ref{sec:results}, we presented a set of network characteristics for graphs obtained from the multi-agent LLM simulation. For the coach evaluation, we primarily relied on reward maximization and average network statistics as evaluation metrics for the tie mechanism. Here, we further examine the degree distributions of the final undirected graphs $\mathcal{G}_T$ after applying the conversion threshold $\theta$. Figure~\ref{fig:degree-distribution-tip-false-heuristic} shows the degree distributions under the heuristic tie mechanism without the coach in the planning process, whereas Fig.~\ref{fig:degree-distribution-tip-true-heuristic} presents the same mechanism with the coach. Likewise, Fig.~\ref{fig:degree-distribution-tip-false-text} reports the distributions under the LLM text-based tie mechanism without the coach, and Fig.~\ref{fig:degree-distribution-tip-true-text} with the coach in the tie reweighting process. The results indicate that the presence of the coach increases the median degree $\tilde{k}$ of the resulting networks, suggesting that coaching induces agents to maintain more active connections on average. The text-based tie formation mechanism yields consistently higher median degrees than the heuristic mechanism across all threshold values, indicating denser and more robust network structures.

% \clearpage

\end{document}